\documentclass[10pt,twocolumn,letterpaper]{article}

\usepackage{iccv}
\usepackage{times}
\usepackage{epsfig}
\usepackage{graphicx}
\usepackage{amsmath}
\usepackage{amssymb}
\usepackage{subcaption}

\usepackage[breaklinks=true,bookmarks=false]{hyperref}

\iccvfinalcopy 

\def\iccvPaperID{2029} 

\begin{document}

\newcommand{\vncomment}[1]{\textcolor{blue}{\bf \small #1 --VN}}
\newcommand{\pccomment}[1]{\textcolor{red}{\bf \small #1 --PC}}

\title{Learning to Estimate Pose by Watching Videos}

\author{Prabuddha Chakraborty and Vinay P. Namboodiri\\
Department of Computer Science and Engineering\\
IIT Kanpur\\
{\tt\small \{prabudc, vinaypn\} @iitk.ac.in}
}

\maketitle

\begin{abstract}
   In this paper we propose a  technique for obtaining coarse pose estimation of humans in an image that does not require any manual supervision. While a general unsupervised technique would fail to estimate human pose, we suggest that sufficient information about coarse pose can be obtained by observing human motion in multiple frames. Specifically, we consider obtaining surrogate supervision through videos as a means for obtaining motion based grouping cues. We supplement the method using a basic object detector that detects persons. With just these components we obtain a rough estimate of the human pose. 

With these samples for training, we train a fully convolutional neural network (FCNN)\cite{long_shelhamer_fcn} to obtain accurate dense blob based pose estimation. We show that the results obtained are close to the ground-truth and to the results obtained using a fully supervised convolutional pose estimation method \cite{cpm} as evaluated on a challenging dataset~\cite{jhmdb}. This is further validated by evaluating the obtained poses using a pose based action recognition method \cite{cheronICCV15}. In this setting we outperform the results as obtained using the baseline method that uses a fully supervised pose estimation algorithm and is competitive with a new baseline created using convolutional pose estimation with full supervision. 
\end{abstract}

\section{Introduction}
\begin{figure}[!ht]
\centering
\begin{subfigure}{.23\textwidth}
\centering
  \includegraphics[height = 1.65in]{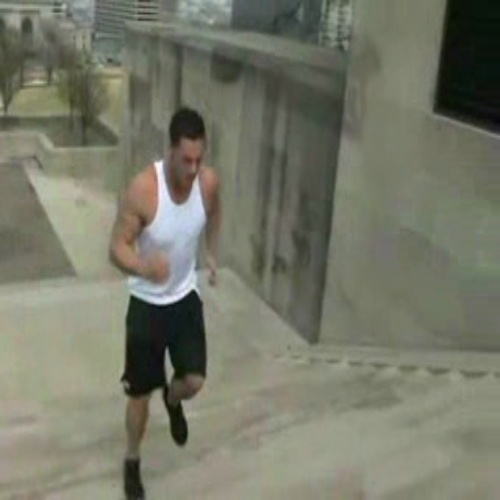}
  \caption{}
  \label{fig:sub1}
\end{subfigure}%
\hfill
\begin{subfigure}{.23\textwidth}
\centering
  \includegraphics[height=1.65in]{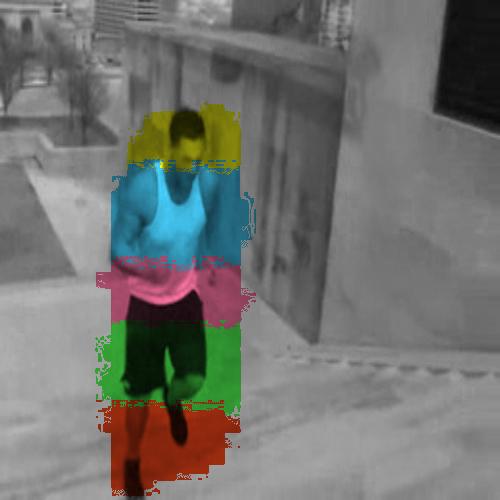}
  \caption{}
  \label{fig:sub2}
\end{subfigure}
\begin{subfigure}{.23\textwidth}
\centering
  \includegraphics[height = 1.65in]{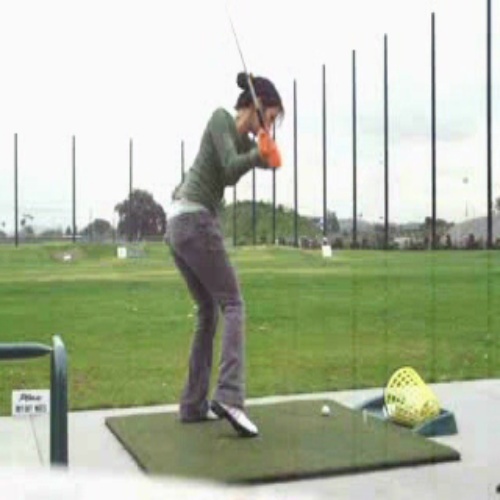}
  \caption{}
  \label{fig:sub1}
\end{subfigure}%
\hfill
\begin{subfigure}{.23\textwidth}
\centering
  \includegraphics[height=1.65in]{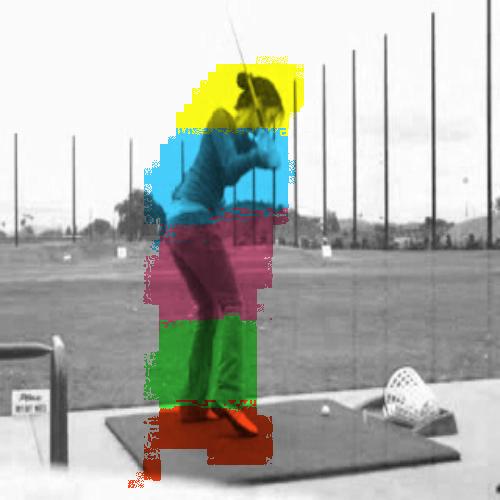}
  \caption{}
  \label{fig:sub2}
\end{subfigure}
\caption{Illustration of pose estimation: Figures (a) and (c) show the original images and (b) and (d) show the respective pose estimates with the various colours depicting the different body parts estimated}
\label{fig:teaser}
\end{figure}

Understanding human pose is a long standing requirement with interesting applications (gaming and other applications using Kinect, robotics, understanding pedestrian behavior, etc.). There has been strong progress over the years particularly using deep learning based pose estimation methods. However, progress is still required for accurate pose estimation in  real world settings. One drawback faced is that the pose estimation methods require manual supervision with explicit labeling of the joint positions. This is particularly more for training state of the art deep learning systems. We address this requirement by proposing a method for obtaining automatic coarse human pose estimates. The method provides us with dense blob based pose estimates that suffices for most practical purposes (such as action recognition). Moreover it is obtained {\it without any manual supervision}. In fig.~\ref{fig:teaser} we illustrate the dense pixel-wise estimates of body parts that are obtained from our method. We can clearly delineate separate regions such as head, neck, torso, knee area and legs as obtained by our method. The use of dense pixel-wise pose estimation allows our method to be robust to a wide variety of pose variations and problems such as occlusions and missing body parts. Further, these are obtained by using only motion cues for the various parts in videos.

The approach in this paper relies on self-supervision or surrogate supervision. Some approaches based on this rely on surrogate tasks such as re-assembling dislocated patches \cite{doersch2015unsupervised} or tracking people \cite{wang2015videorep}. 
An interesting recent line of work that is related to this work relies on learning segmentation by using motion flows \cite{pathak2016motion}. These surrogate tasks can be used for obtaining visual representations for generic tasks like classification or segmentation. Visual representations obtained through the techniques proposed so far however do not address granular tasks such as human pose estimation. Yet, we as humans can solve the problem easily. The primal cue that enables us in this task is observing the motion of the different body parts. This was evident early on and used by Gunnar Johansson in his seminal early work that analysed human body motion \cite{johansson1973}. In this work Johansson observed that the relative motion between the body parts can be used for analysing human pose. Inspired by this insight we use the relative grouping of motion flow of humans to obtain the pose supervision required.  

Our approach uses embarrassingly simple techniques that can be easily obtained in any setting for obtaining automatic supervision. These can always be improved upon. Our aim in using these techniques was to show that even the most basic grouping of human motion flow suffices to obtain the supervision required to be competitive to current state of the art techniques trained using carefully annotated supervised data. Interestingly, with enough data, the deep network learns to generate output parts that are substantially better than the noisy supervision provided as input. The results are evaluated in terms of pose estimate comparisons as well as components of an action recognition method. The end-result is a competitive pose estimation method for free (zero supervision cost) by using easily available video data.



\section{Related Work}
\label{sec:lit_survey}
There are two streams of work that are of relevance to the proposed work:
\subsection{Pose Estimation}
Human pose estimation has been solved by estimating a deformable mixture of body parts by Felzenszwalb and Huttenlocher \cite{felzenszwalb_ijcv2005}. This method provides a robust estimation of pose by using a spring and parts model allowing for deformation of human pose. The human body deformation is a significant challenge in pose estimation and this line of work allows for such deformation. This line of work has been successfully followed up by Andriluka {\it et al.} \cite{Andriluka:2009} and Eichner {\it et al.} \cite{Eichner12}. Johnson and Everingham \cite{johnson_cvpr11} consider a method that is able to learn from inaccurate annotation. In their work, the authors use clustered centroids obtained by a larger dataset to obtain cluster specific priors for pose estimation. While, this approach is pertinent to our aim of working in the presence of noisy annotation, we are able to tolerate much larger inaccuracies than is considered in this work. Ladicky {\it et al.} \cite{Ladicky13} consider an interesting approach that combines pixel wise pose estimation with pictorial structures based pose estimation. In our work, we consider only pixel wise pose estimation. The advantage is that this pose estimation is more tolerant to occlusion of joints in various poses. We observe this phenomenon that pixel wise pose estimation similarly provides robustness towards occlusion and missing body parts. Ramakrishna {\it et al.} \cite{Ramakrishna_2014} in their work move beyond tree structured models by using inference machines that allow for richer interaction and better estimates of the parts by considering joint structured output prediction inference machines. While, similar in nature, we use recent advances in deep learning to avoid explicit structured representation learning by allowing fully convolutional networks to provide data dependent prediction. An related line of work is the seminal work by Shotton {\it et al.} \cite{shotton} where the authors used synthetic renderings in order to estimate pose in depth images. This work however, is applicable to depth images and not to real-world color images. 

Recently there have been a number of approaches that target solving the pose estimation problem in the deep learning framework \cite{toshev, cpm, jain_iclr14}. An initial deep learning based approach was proposed by Jain {\it et al.} \cite{jain_iclr14} where the authors considered a number of independent convolutional neural networks used for binary prediction of each body part. This binary classifier was used in a sliding window approach to generate a response map per body part. Subsequent work from Toshev and Szegedy \cite{toshev} follow an interesting pose estimation approach that uses a cascade of deep regressors for pose estimation.  At the first stage the architecture predicts the initial pose with the subsequent stages predicting finer pose in terms of displacement from the initial predicted pose. This approach of using sequential prediction is also adopted by Wei {\it et al.} \cite{cpm} in their work that allows for sequential prediction in multiple stages with each stage operating on the belief map of the previous stage. In our work, we adopt the fully convolutional segmentation prediction framework \cite{long_shelhamer_fcn} that is easier to train. Further, none of the methods so far considered could be trained without requiring manual supervision for training. As is well known by the community, each training set has its own bias and methods trained in one scenario would not work well in other scenarios due to a domain shift or dataset bias \cite{Efros_2011_6976}. Our approach due to its ability to automatically generate supervision for training could always be applied in any novel scenario by just obtaining relevant data and obtaining automatic supervision through simple methods.

\subsection{Self-supervision}
There have been a number of works that are based on self-supervision or surrogate supervision. The initial methods were aimed at obtaining unsupervised means of generating visual representations that were competitive to supervised object classification task by performing other tasks for which supervision was directly obtainable such as context prediction \cite{doersch2015unsupervised} or ego-motion \cite{jayaraman-iccv2015} or by tracking objects in videos \cite{wang2015videorep}. Subsequently this concept has been explored for a wide range of tasks such as learning visual representations by using robotic motions \cite{LSM2015, Pinto2016}. Further recent works include using the task of inpainting an image \cite{pathakCVPR16context} or predicting the odd subsequence from a set of video sub-sequences \cite{Fernando2017}. The task of self-supervision for a semantically granular task such as human pose estimation has not yet been solved by the methods proposed so far. In the next section we provide details of the proposed method for obtaining self supervision for solving the problem of pose estimation.

\begin{figure}[t]
  \centering
  \includegraphics[width=3in]{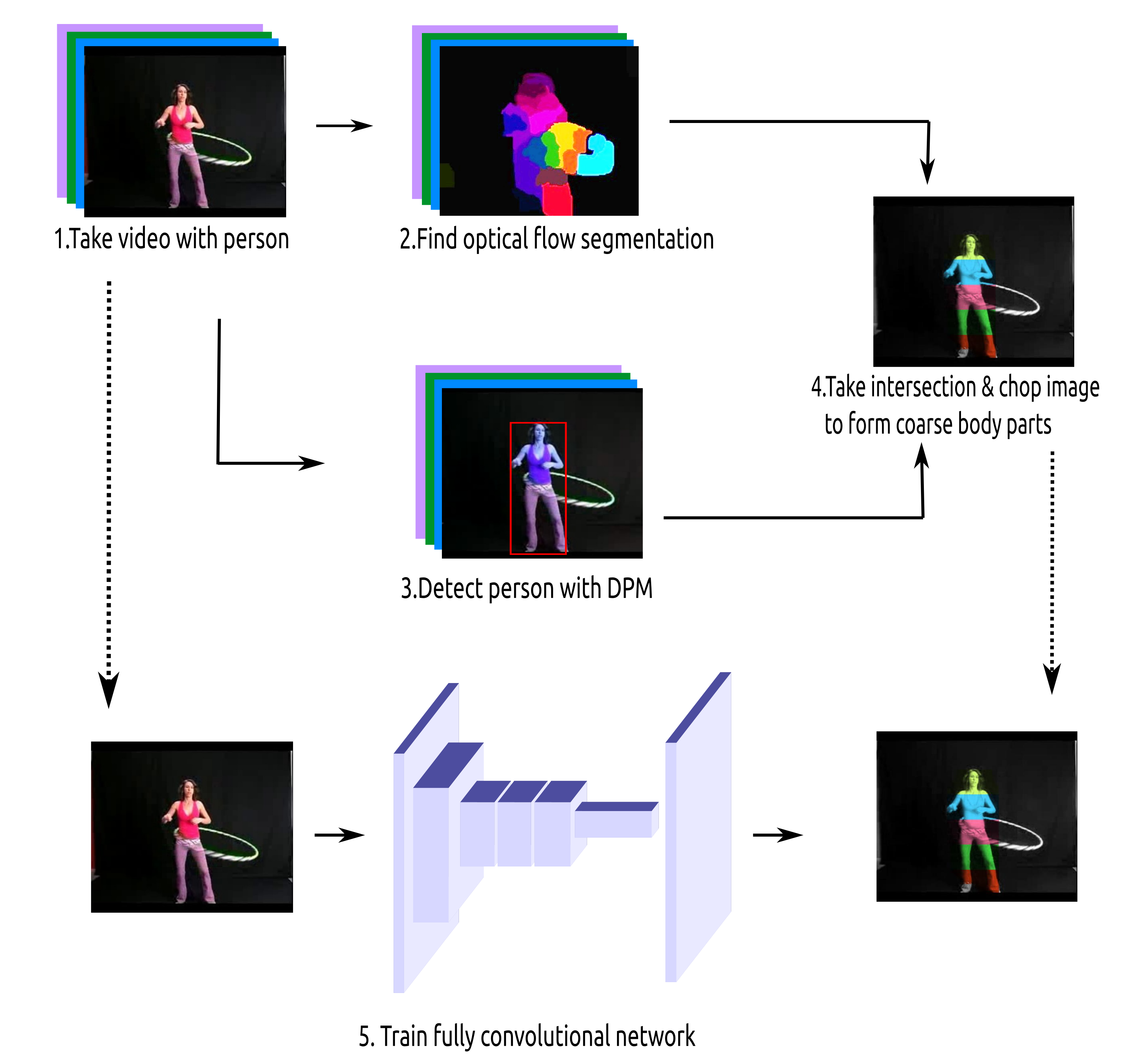}
  \caption{Illustration of the method}
  \label{fig:pipeline}
\end{figure}

\section{Method}
\label{sec:method}
Our method is a simple sequence of steps that provides the coarse supervision necessary for pose estimation as illustrated in figure~\ref{fig:pipeline}. We obtain the dataset in terms of videos with very little assumptions on the videos. We have evaluated using videos from two action recognition datasets for obtaining training data, viz. UPenn action recognition dataset \cite{penn_ds} and UCF 101 action dataset \cite{ucf_101}.  We obtain optical flows between consecutive pairwise frames in a video from the videos in a dataset using Farneback's optical flow technique \cite{farneback}. We use two thresholds on the flow, one to ensure that there is some motion in the frame (more than 10\% pixels are having optical flow values above zero) and the other to ensure that the whole frame is not moving (less than 70\% of the frame has optical flow values above zero). Using this motion flow we group the optical flow values into blobs using a simple mean shift based grouping technique \cite{mean_shift}. This step yields blobs of motion flow that are grouped. We then need to ensure that the motion flow contains motion from a person and not some extraneous source such as motion of vehicles or other moving objects such as swings or animals. This is done by using a deformable part model based person detector \cite{dpm}. We observed that the root filter predictions could be used to prune non-person blobs from person-blobs. These were noisy detections (as shown in section 4.6), however, as can be seen from the experimental section, these proved sufficient for obtaining reasonable training supervision.

From the sequence of steps above, we obtain a set of frames that have person detections and motion flow blobs. The intersection of these two steps is used to obtain a set of blobs for detected persons. In our method we use only the frames with a single person detection per frame as training data. This simplifying assumption allows us to avoid the problem of forming the association of motion flow blobs to multiple persons during training. The method learned is able to estimate a set of motion flow blob segments that belongs to a single person. Note that this does not limit our method and multiple persons pose estimation can be predicted during testing as is shown in figure \ref{fig:results}(s). Having obtained the blobs for a person, we now have to obtain the part estimates. In our method, we divide the root filter horizontally into five parts that coarsely provides pose estimates corresponding to head, torso and arms, and legs. These are obtained by uniformly dividing the root filter detection bounding box into five equal horizontal blocks. The resulting bounding boxes result in coarse pose estimation that still corresponds rather accurately to the five parts as is verified experimentally. We evaluated various number of horizontal bands (discussed in section 4.5) and observed that five parts was providing us with an appropriate number of parts   that was discriminative and  representative of the human pose as required for recognizing actions.

Having obtained these coarse pose supervision, we train a fully convolutional neural network for segmentation \cite{long_shelhamer_fcn} that we adapt for segmenting pose estimation blobs. The whole pipeline is illustrated in fig.~\ref{fig:pipeline} where we show how videos are used to obtain optical flow that is segmented using mean shift to provide motion blobs. Further the DPM based detector is used to provide person detections. The intersection of the motion blobs with the person detections provides us with estimates of the parts of a moving person. These are divided into five horizontal partitions resulting in five dense pixel-wise part estimates. These are then trained using a fully convolutional neural network (FCNN) \cite{long_shelhamer_fcn} to generate pixel-wise estimates of the five part segments. Each of the steps in our pipeline (except for the final segmentation prediction step) uses basic building blocks and can be improved upon. The main aim was to ensure that our method is not contingent on an advanced building block and even the simplest of building blocks suffices to obtain automatic supervision for pose estimation. In the next section we evaluate this basic approach thoroughly and compare it competitively with state of the art pose estimation techniques.

\section{Experimental Evaluation}
\label{sec:experiments}
In this section we initially describe the experimental setup, followed by a quantitative evaluation of body pose estimates. We then use the pose estimate as a component for action recognition and provide a comparison. Next, we consider the effect of amount of data and number of parts followed by qualitatively considering the results for object localizatione and visualising the results for a number of samples.

\subsection{Experimental Setup}

\textbf{Dataset}  For training the fully convolution neural network we have used videos from UCF-101 \cite{ucf_101} and Penn Action Dataset \cite{penn_ds}. 

\textbf{Training} 
We trained the network with a minibatch size of 10 using adam optimizer. For training the model with 40k images we used a learning rate $10^{-4}$,  beta1 0.9 , beta2 0.999 and no decay. 

All our models are implemented with Keras having Theano backend using NVIDIA GeForce GTX TITAN X. Further details regarding the method is available in our public repository \footnote{\textcolor{red}{\url{https://github.com/prabuddha1/acpe/} } }

\textbf{Hard Mining} 
After we obtained the model trained with 20,000 images, we trained it further on 20,000 more images, sampled from 60,000 images, for which our model was inaccurate. We could thus reduce the number of images we needed to consider. This provided us with our final model that was trained with a total 40,000 images.  

\subsection{Body Pose Estimate comparison}
We compare our proposed method for pose estimation against convolutional pose machine (CPM)~\cite{cpm} method that is the best model present trained using the Leeds dataset \cite{LSP} and MPII pose dataset \cite{mpii}. We obtain distance of the part locations from the ground-truth part locations in JHMDB dataset \cite{jhmdb}. The exact part locations are obtained as centroids of the parts for our method whereas they are directly predicted using CPM~\cite{cpm}. As can be observed from the results presented in table~\ref{table:pose_data}, for various part locations the results are quite close to the ground-truth part locations on average. The predictions are especially better as compared to CPM for part 5 that predicts the part around knees. As the distance from the torso increases it becomes harder to predict and so this part is a difficult part to reliably predict. The other parts such as the part around face and belly are also very close. The part around hips and shoulders are harder as they are not consistently obtained through our automatic annotation. The results for the automatic pose generation method is definitely much worse as compared to the output obtained after training. Note that our method is not trained on JHMDB, but only on UPenn and UCF datasets without using any pose ground-truth. The performance gap is clearly visible by considering the distances obtained in the second column against those obtained by our method in the third column. This is also evident in section 4.6 when we consider the object localisation results as the outputs obtained by the DPM detector \cite{dpm} are qualitatively much worse as compared to the localisation we obtain. We further evaluate our method on a subset of MPII pose dataset with 17372 training images. For this we use the best CPM model not trained using MPII dataset as the training images are used and test it with our model trained on 40,000 images from UCF and Penn datasets. In this setting we observe that we are able to outperform CPM in most of the part estimates as shown in table~\ref{table:pose_data_mpii}

\begin{table}[h]
\centering
\begin{tabular}{ |p{1.7cm}||p{1.7cm}| |p{1.7cm}| |p{1.7cm}| }
 \hline
 \multicolumn{4}{|c|}{Comparison of Pose estimation on JHMDB dataset \cite{jhmdb} } \\
 \hline
Part Name  & Distance of CPM from ground truth~\cite{cpm} & Distance of Pose Supervision Generator- & Distance with our 40k Image Model trained on Penn~\cite{penn_ds} and UCF 101~\cite{ucf_101}\\
\hline
 \multicolumn{4}{|c|}{Average Euclidean Distance 1 unit = 1 pixel}\\
 \hline
  Face - Part-1 & 38.93 & 58.11 & 40.46\\
   \hline
  Between Shoulders - Part-2 & 27.47 & 55.08 & 39.82\\
   \hline
  Belly - Part-3 & 55.10 & 68.60 & 55.76\\
   \hline
  Between Hips -- Part-4 & 50.54 & 70.87& 61.72\\
   \hline
  Between Knees -- Part-5 & 87.11 & 88.45& 77.38\\
  \hline
  Between Ankles -- Part-5 & 112.09 & 116.54& 92.0088\\
 \hline
\end{tabular}
\caption{In this table we provide a comparison of pose estimates with the ground-truth pose in JHMDB dataset. The CPM ~\cite{cpm} model is trained with MPII \cite{mpii} and LSP \cite{LSP} datasets. Our method is trained with automatic annotation on other videos (not JHMDB) without manual supervision.}
\label{table:pose_data}
\end{table}

\begin{table}[h]
\centering
\begin{tabular}{ |p{1.7cm}||p{1.7cm}|  |p{1.7cm}| }
 \hline
 \multicolumn{3}{|c|}{Comparison of Pose estimationon MPII dataset \cite{mpii}} \\
 \hline
Part Name  & Distance of CPM ~\cite{cpm}  & Distance with our model\\
\hline
 \multicolumn{3}{|c|}{Average Euclidean Distance 1 unit = 1 pixel}\\
 \hline
 Part-1 & 214.05  & 209.55\\
   \hline
Part-2 & 210.14  & 183.63\\
   \hline
Part-3 & 285.14  & 245.16\\
   \hline
Part-4 & 291.01 & 255.48 \\
   \hline
Part-5 (knees) & 369.46 & 393.66\\
   \hline
Part-5 (Ankles) & 428.36 & 462.70\\
 \hline
\end{tabular}
\caption{In this table we provide a comparison of pose estimates with the ground-truth pose in MPII  dataset \cite{mpii}. This test is performed on 17372 training images from MPII training component. The CPM model used is trained on LSP\cite{LSP} }
\label{table:pose_data_mpii}
\end{table}

\subsection{Pose estimation in Action Recognition}
We next evaluate our method indirectly by considering its use in action recognition. We do this through an action recognition method that uses pose for recognizing action  proposed by Cheron {\it et al.}~\cite{cheronICCV15}. Their method uses a supervised pose estimation method~\cite{cherian} that they had proposed earlier that especially handles mixed body poses. The actions are evaluated on the realistic JHMDB dataset~\cite{jhmdb}. We compare the action recognition accuracy by also considering the state-of-the-art CPM pose estimation method that is the best model present trained using the Leeds dataset and MPII pose dataset. This is not a fair comparison as our method is not trained with manual supervision. However, as can be observed from the results shown in table~\ref{table:result}, we out-perform the supervised method of P-CNN~\cite{cheronICCV15} using mixed body pose estimates~\cite{cherian} even in this setting by around 2.2\%. Small improvements can be obtained by varying the PCNN parameters improving the accuracy of our method to around 65.01\%, but as this would not be the result of the pose estimation, but rather the recognition method, we do not consider such optimizations in the rest of the paper and report the original value obtained in the table~\ref{table:result}. Thus, our method does not attain the accuracy of P-CNN with CPM features, however, we are close to their performance and the proposed method can be further improved by validating the pose estimation with P-CNN method parameters or fine-tuning on the JHMDB dataset. Such optimisations are not currently considered in our method. 

\begin{table}[h!]
\centering
\begin{tabular}{ |p{3cm}||p{2cm}|  }
 \hline
 \multicolumn{2}{|c|}{Action recognition using P-CNN~\cite{cheronICCV15} }\\
\multicolumn{2}{|c|}{A comparison with various pose estimation methods}\\
 \hline
Method Name     & Accuracy \\
 \hline
  Mixed body pose~\cite{cherian}&   61.1\%\\
 CPM~\cite{cpm}   & 66.13\%\\
 Proposed Method & 63.26\%\\
 \hline
\end{tabular}
\caption{In this table we provide a comparison of various pose estimation methods evaluated through recognition of actions using the JHMDB dataset. The proposed method is still competitive  even in the absence of ground truth training}
\label{table:result}
\end{table}

\subsection{Varying amount of data}
\begin{figure}[h]
  \centering
  \includegraphics[width=2.8in]{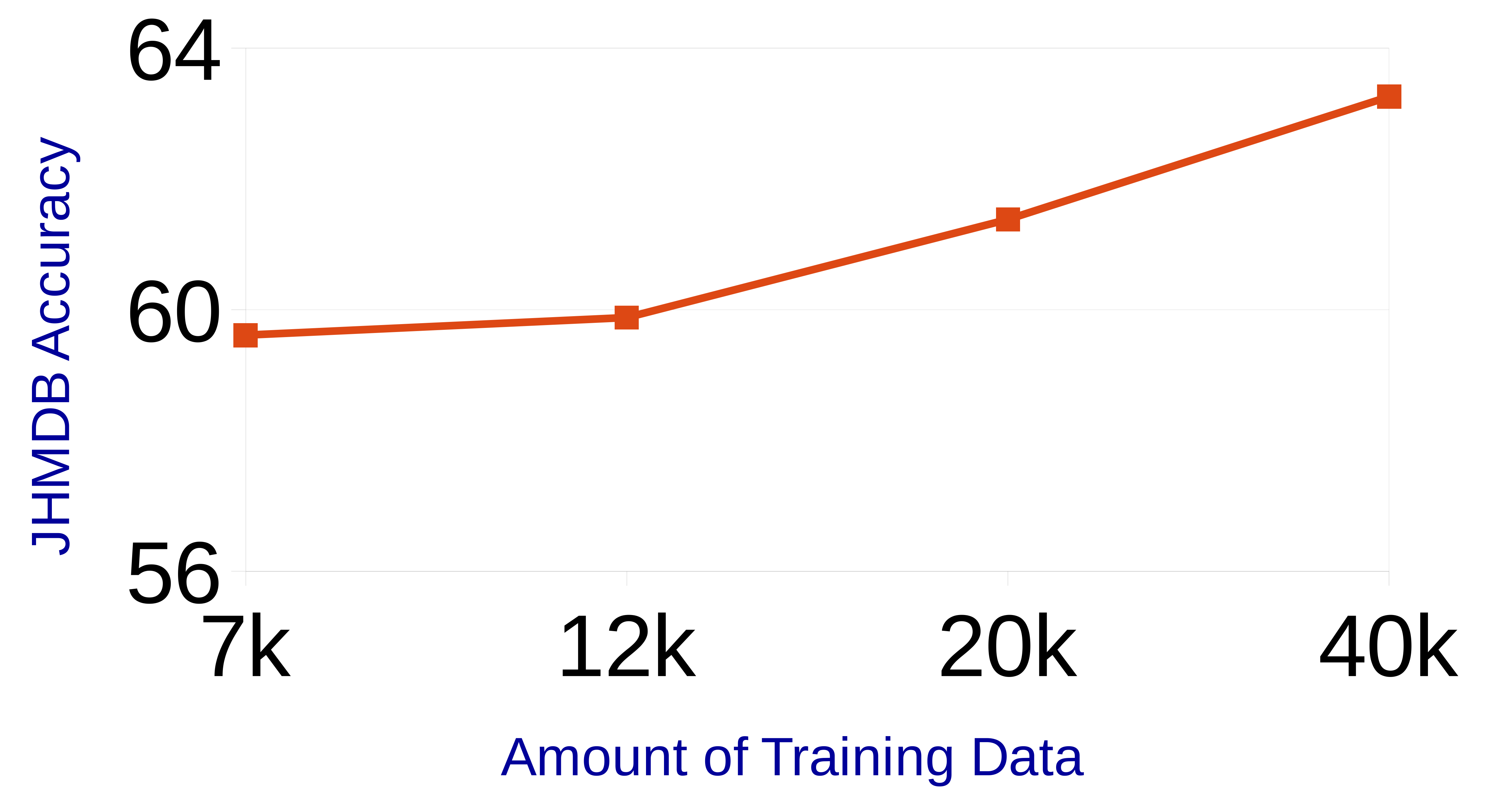}
  \caption{Difference in accuracy as the number of data samples is increased from 7000 to 40,000 samples. Increasing amount of data continuously increases the performance of the method.}
  \label{fig:data}
\end{figure}

\begin{figure*}
\centering
\begin{subfigure}{.19\textwidth}
\centering
  \includegraphics[height = 1.38in]{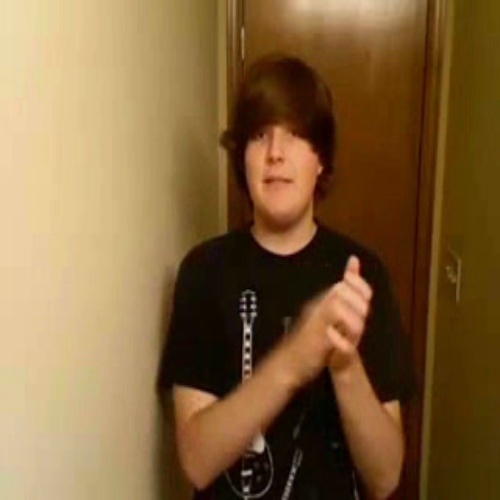}
  \caption{}
  \label{fig:sub1}
\end{subfigure}%
\hfill
\begin{subfigure}{.19\textwidth}
\centering
  \includegraphics[height=1.38in]{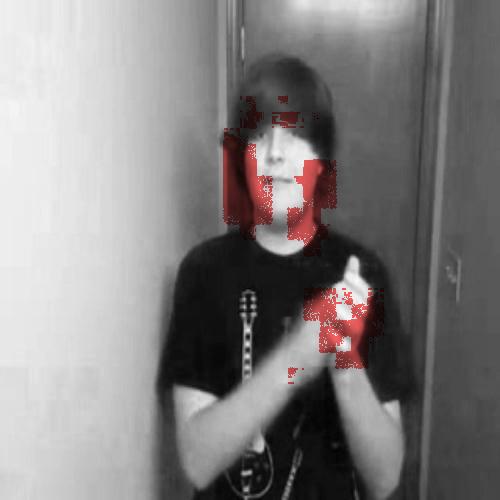}
  \caption{}
  \label{fig:sub2}
\end{subfigure}%
\hfill
\begin{subfigure}{.19\textwidth}
\centering
  \includegraphics[height = 1.38in]{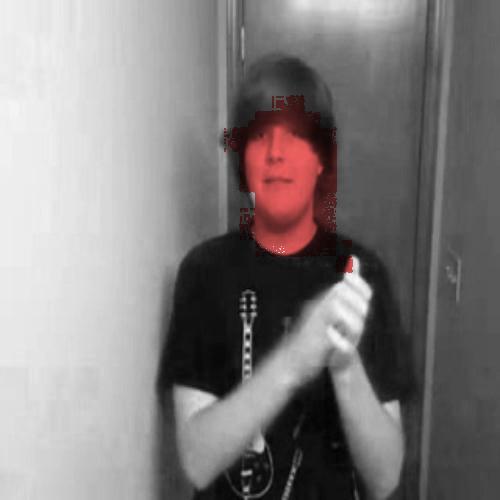}
  \caption{}
  \label{fig:sub1}
\end{subfigure}%
\hfill
\begin{subfigure}{.19\textwidth}
\centering
  \includegraphics[height=1.38in]{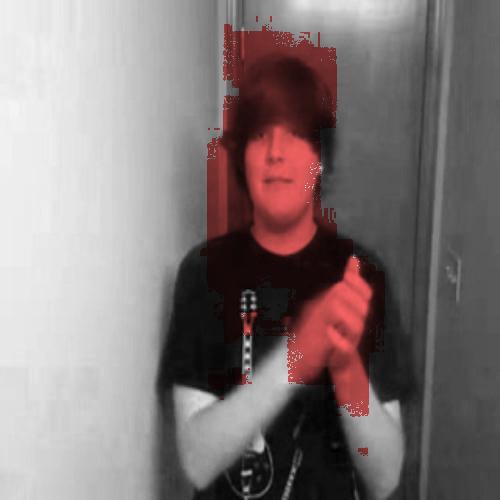}
  \caption{}
  \label{fig:sub2}
\end{subfigure}%
\hfill
\begin{subfigure}{.19\textwidth}
\centering
  \includegraphics[height=1.38in]{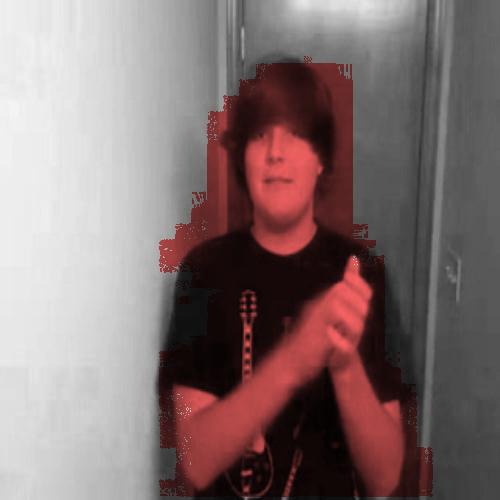}
  \caption{}
  \label{fig:sub2}
\end{subfigure}

\centering
\begin{subfigure}{.19\textwidth}
\centering
  \includegraphics[height = 1.38in]{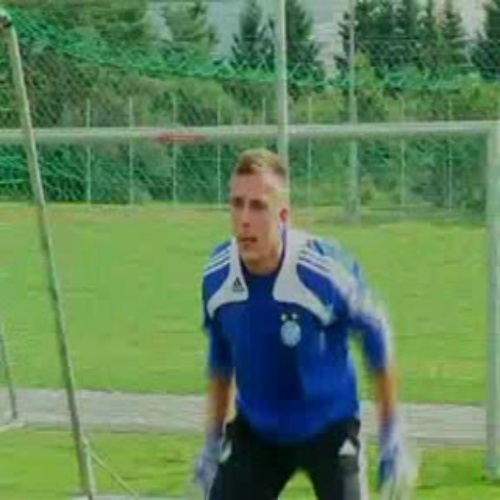}
  \caption{}
  \label{fig:sub1}
\end{subfigure}%
\hfill
\begin{subfigure}{.19\textwidth}
\centering
  \includegraphics[height=1.38in]{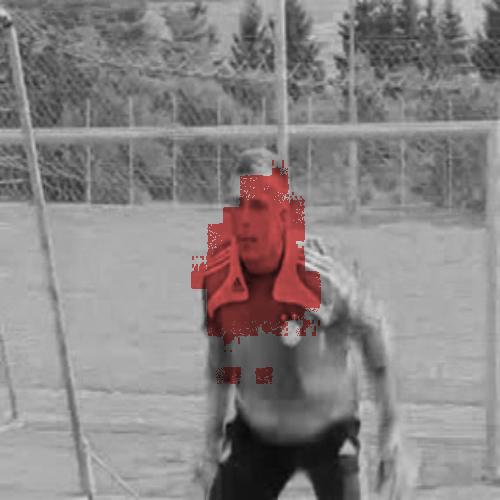}
  \caption{}
  \label{fig:sub2}
\end{subfigure}%
\hfill
\begin{subfigure}{.19\textwidth}
\centering
  \includegraphics[height = 1.38in]{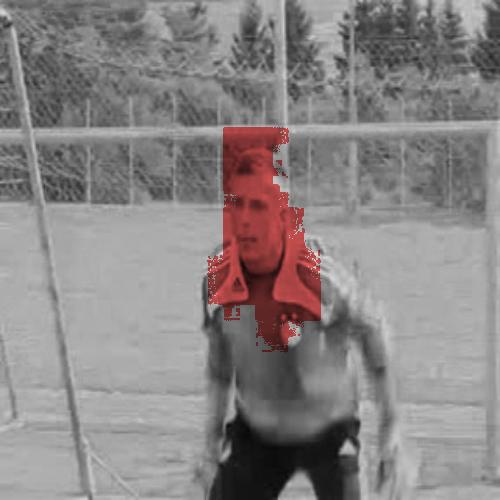}
  \caption{}
  \label{fig:sub1}
\end{subfigure}%
\hfill
\begin{subfigure}{.19\textwidth}
\centering
  \includegraphics[height=1.38in]{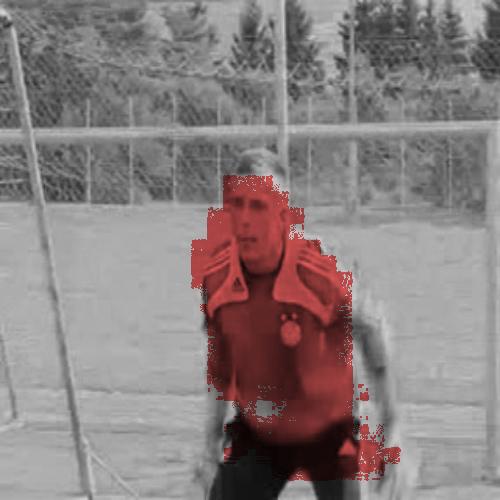}
  \caption{}
  \label{fig:sub2}
\end{subfigure}%
\hfill
\begin{subfigure}{.19\textwidth}
\centering
  \includegraphics[height=1.38in]{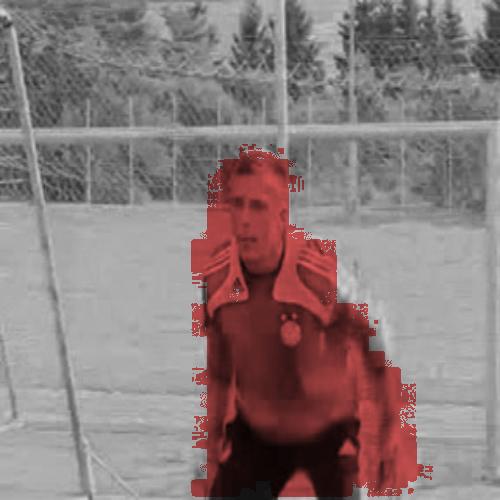}
  \caption{}
  \label{fig:sub2}
\end{subfigure}
  \caption{Difference in accuracy in full body estimation against the amount of data. The amount of data used for training is from left to right 7000 samples, 12000 samples, 20000 samples and 40000 samples.}
  \label{fig:data_illus}
\end{figure*}

We next evaluate our method using the action recognition setting to analyse how the amount of data would affect the result. The results are illustrated in the graph shown in figure~\ref{fig:data}. As can be observed from the graph, the results consistently improved. The amount of data-samples used for training the fully convolutional neural network through automatic annotation is varied from 7000 samples to 40,000 samples. The addition of samples has aided the recognition and we were constrained only in terms of physical memory limitations in terms of the data-set with which we could train the system. Normally, any method is usually limited by amount of supervised training data available and this is not a constraint for our method. We can visualize this qualitatively in figure~\ref{fig:data_illus} by observing variation of the result in terms of extraction of all the parts jointly as we increase the data. As can be seen, as we increase the amount of data, the full body extraction of the person is increasingly improved. This is reflected in the results as well as shown in the graph~\ref{fig:data}.

\subsection{Varying number of parts}
\begin{figure}[h]
  \centering
  \includegraphics[width=2.8in]{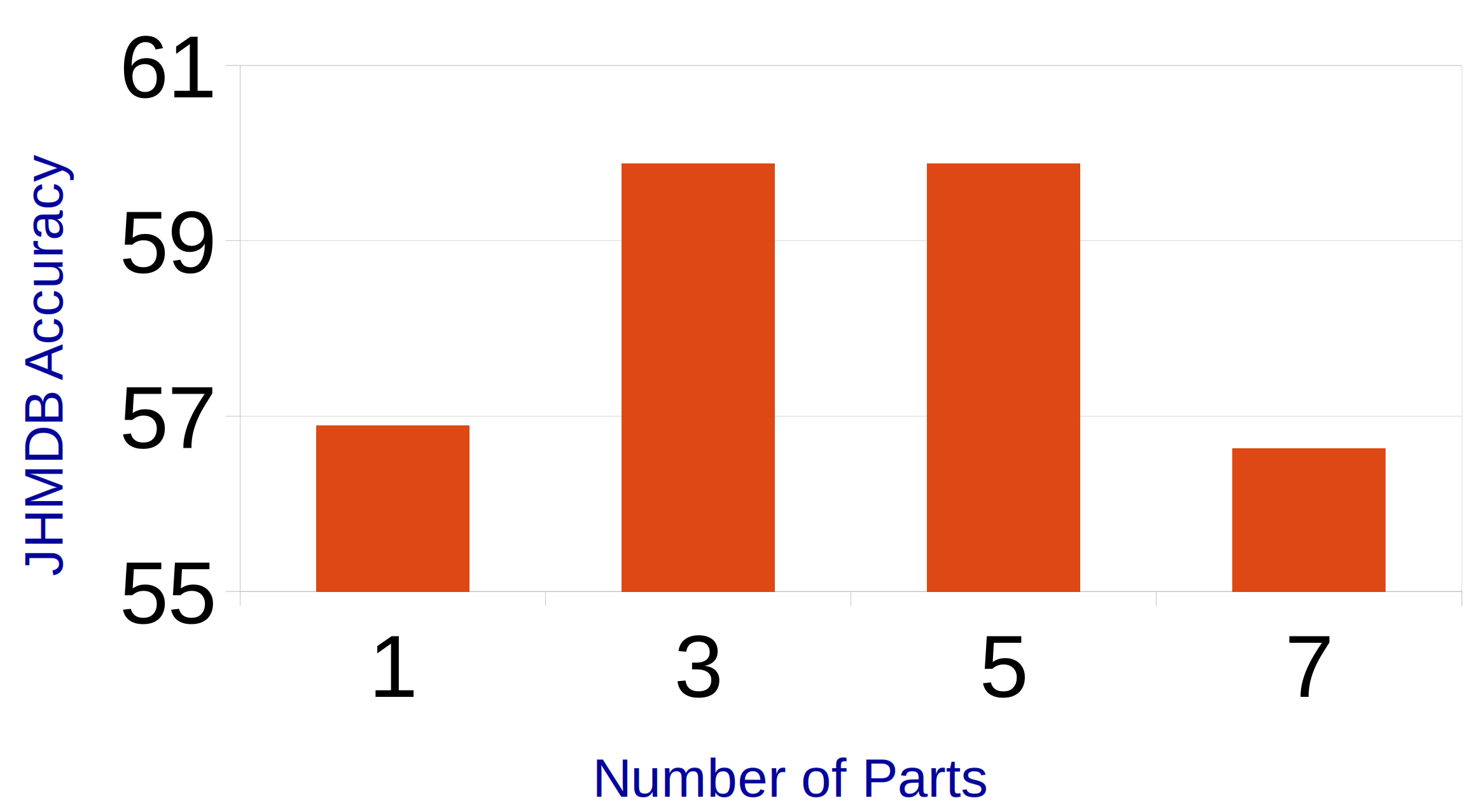}
  \caption{Difference in accuracy in action recognition task against number of parts}
  \label{fig:parts}
\end{figure}

\begin{figure*}
\centering
\begin{subfigure}{.19\textwidth}
\centering
  \includegraphics[height = 1.38in]{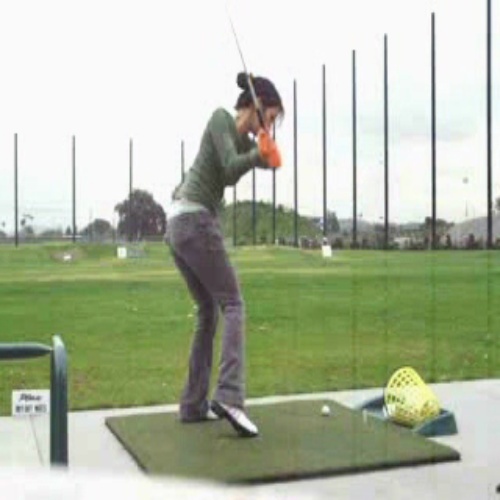}
  \caption{}
  \label{fig:sub1}
\end{subfigure}%
\hfill
\begin{subfigure}{.19\textwidth}
\centering
  \includegraphics[height=1.38in]{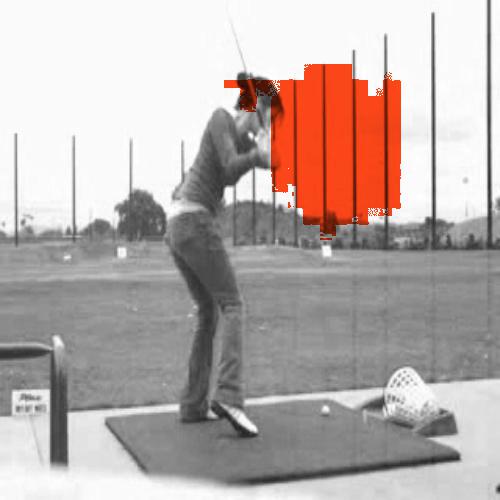}
  \caption{}
  \label{fig:sub2}
\end{subfigure}%
\hfill
\begin{subfigure}{.19\textwidth}
\centering
  \includegraphics[height = 1.38in]{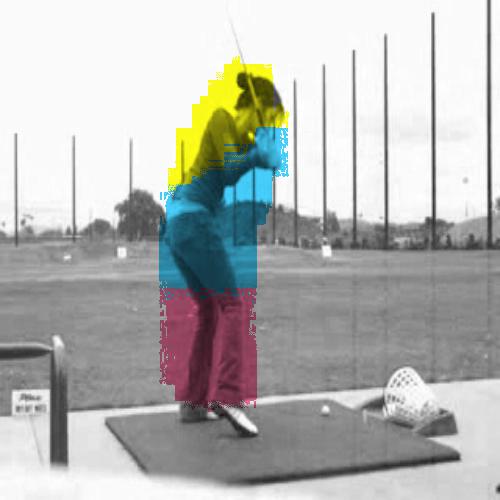}
  \caption{}
  \label{fig:sub1}
\end{subfigure}%
\hfill
\begin{subfigure}{.19\textwidth}
\centering
  \includegraphics[height=1.38in]{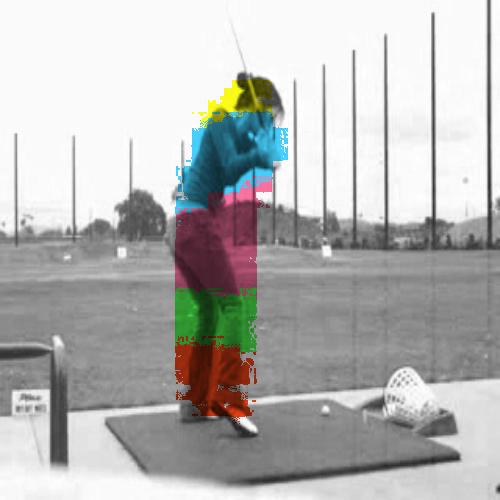}
  \caption{}
  \label{fig:sub2}
\end{subfigure}%
\hfill
\begin{subfigure}{.19\textwidth}
\centering
  \includegraphics[height=1.38in]{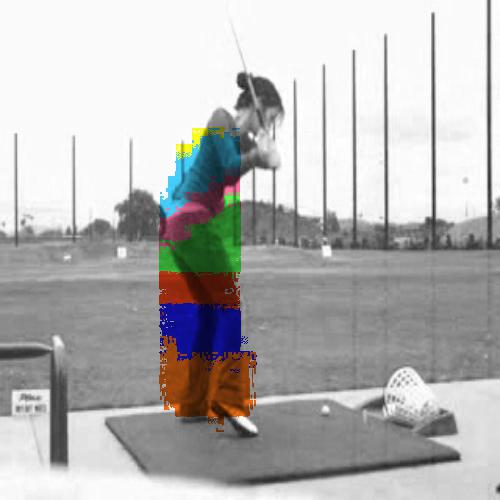}
  \caption{}
  \label{fig:sub2}
\end{subfigure}

\centering
\begin{subfigure}{.19\textwidth}
\centering
  \includegraphics[height = 1.38in]{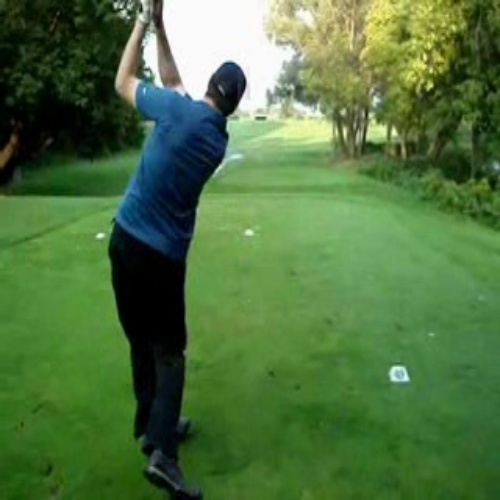}
  \caption{}
  \label{fig:sub1}
\end{subfigure}%
\hfill
\begin{subfigure}{.19\textwidth}
\centering
  \includegraphics[height=1.38in]{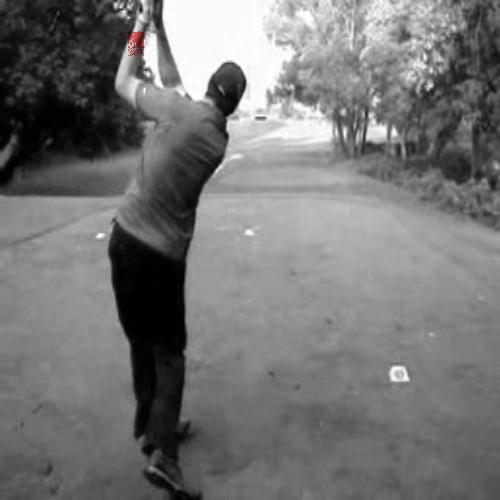}
  \caption{}
  \label{fig:sub2}
\end{subfigure}%
\hfill
\begin{subfigure}{.19\textwidth}
\centering
  \includegraphics[height = 1.38in]{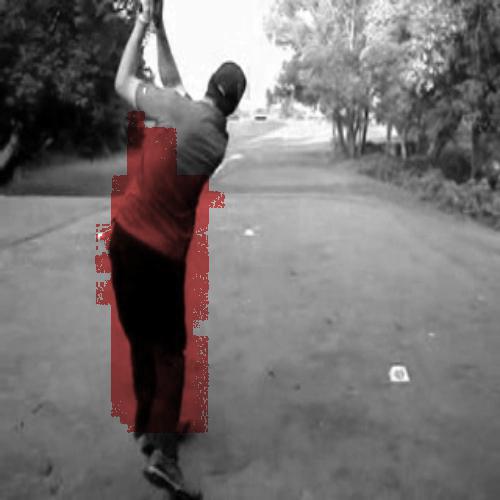}
  \caption{}
  \label{fig:sub1}
\end{subfigure}%
\hfill
\begin{subfigure}{.19\textwidth}
\centering
  \includegraphics[height=1.38in]{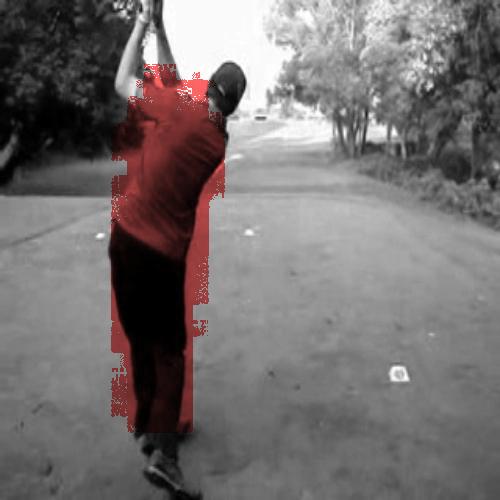}
  \caption{}
  \label{fig:sub2}
\end{subfigure}%
\hfill
\begin{subfigure}{.19\textwidth}
\centering
  \includegraphics[height=1.38in]{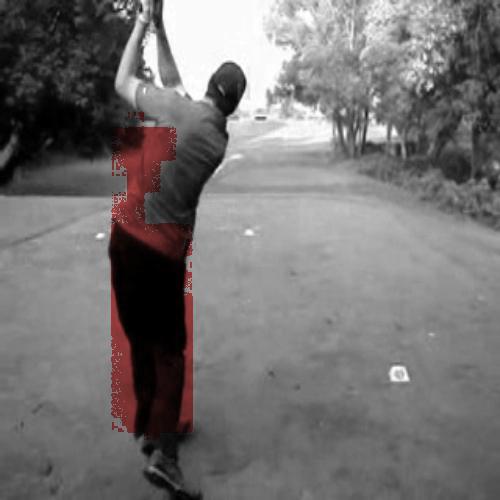}
  \caption{}
  \label{fig:sub2}
\end{subfigure}
  \caption{Difference in pose estimation using proposed method as obtained by varying the number of parts}
  \label{fig:part_illus}
\end{figure*}

We next analyse the effect of number of parts in our proposed method. We evaluate the effect of varying the number of parts for the task of action recognition on the JHMDB dataset. As can be observed from the graph~\ref{fig:parts} we obtained maximum accuracy using 5 parts. This experiment was carried out by fixing the number of samples to around 12000 samples and varying the number of parts. We can also observe this phenomenon visually in figure~\ref{fig:part_illus}. Using a single part we observe in figure~\ref{fig:part_illus} that the pose estimation is attracted towards the golf club as a single part and does not detect the man or woman. With three parts, the pose estimation improves and we obtain three gross parts. This is further improved and tightly obtained when we use five parts. With seven parts, the individual part samples are not discriminative enough and are not reliably estimated. In figure~\ref{fig:part_illus}(f) - (j) we consider the whole body being estimated by considering different number of parts as a slight mismatch in the individual parts may be tolerated. As can be seen the figure~\ref{fig:part_illus}(i) the model with 5 parts provides us the best estimate of the person as a whole as compared to other varying number of parts. We therefore use five parts in our proposed method for all the remaining experiments. 

\subsection{Qualitative results and comparison}
We now obtain the comparison of the proposed method qualitatively with the Faster RCNN \cite{faster_rcnn} that was trained on Pascal VOC using ground-truth data and analyse the results from the proposed method qualitatively. 

\begin{figure}
\centering
\begin{subfigure}{.1\textwidth}
\centering
  \includegraphics[height = 0.85in]{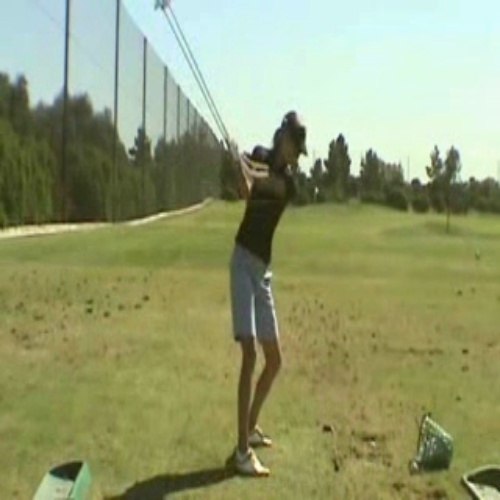}
  \caption{}
  \label{fig:sub1}
\end{subfigure}%
\hfill
\begin{subfigure}{.1\textwidth}
\centering
  \includegraphics[height = 0.85in]{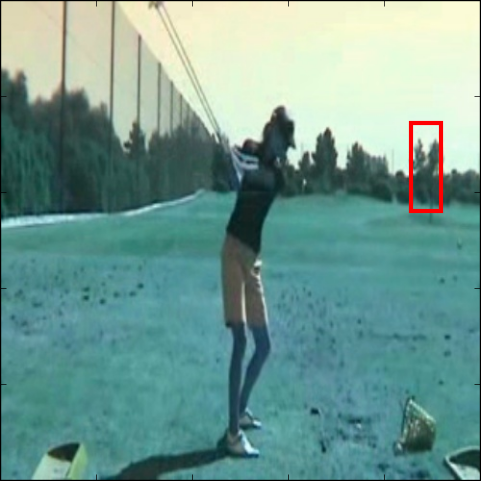}
  \caption{}
  \label{fig:sub1}
\end{subfigure}%
\hfill
\begin{subfigure}{.1\textwidth}
\centering
  \includegraphics[height = 0.85in]{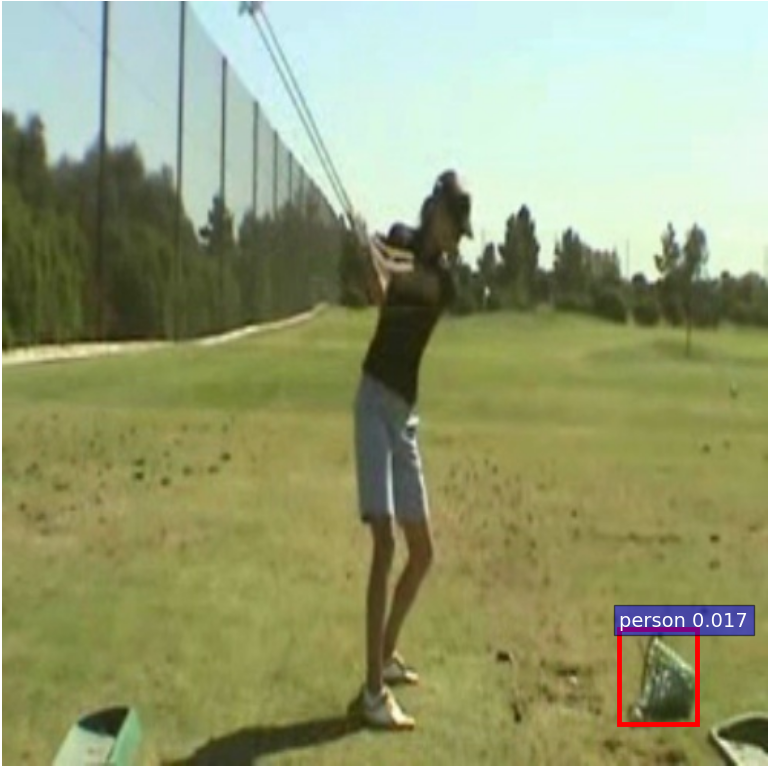}
  \caption{}
  \label{fig:sub1}
\end{subfigure}%
\hfill
\begin{subfigure}{.1\textwidth}
\centering
  \includegraphics[height = 0.85in]{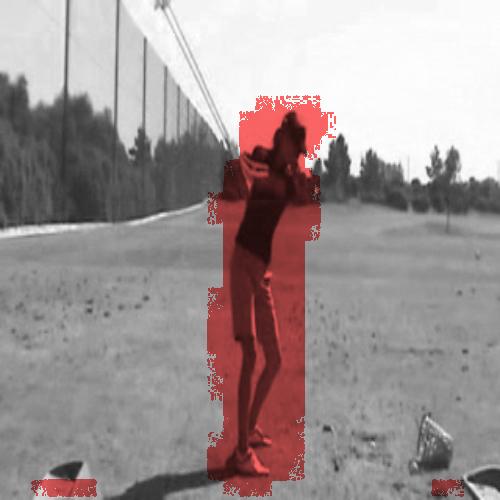}
  \caption{}
  \label{fig:sub1}
\end{subfigure}

\begin{subfigure}{.1\textwidth}
\centering
  \includegraphics[height = 0.85in]{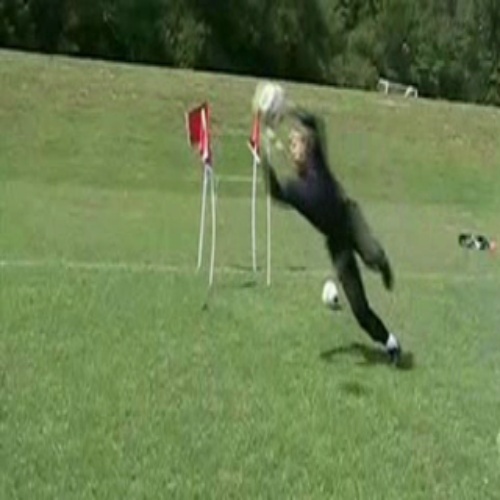}
  \caption{}
  \label{fig:sub1}
\end{subfigure}%
\hfill
\begin{subfigure}{.1\textwidth}
\centering
  \includegraphics[height = 0.85in]{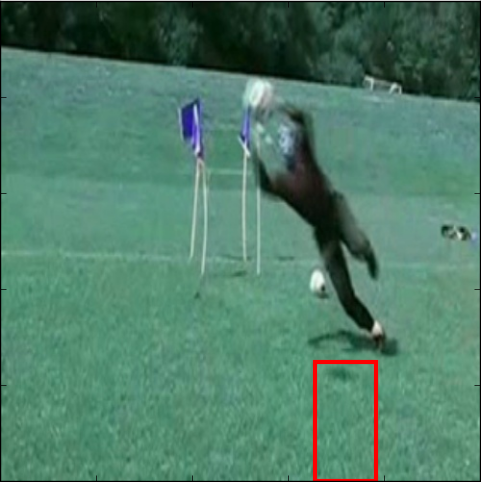}
  \caption{}
  \label{fig:sub1}
\end{subfigure}%
\hfill
\begin{subfigure}{.1\textwidth}
\centering
  \includegraphics[height = 0.85in]{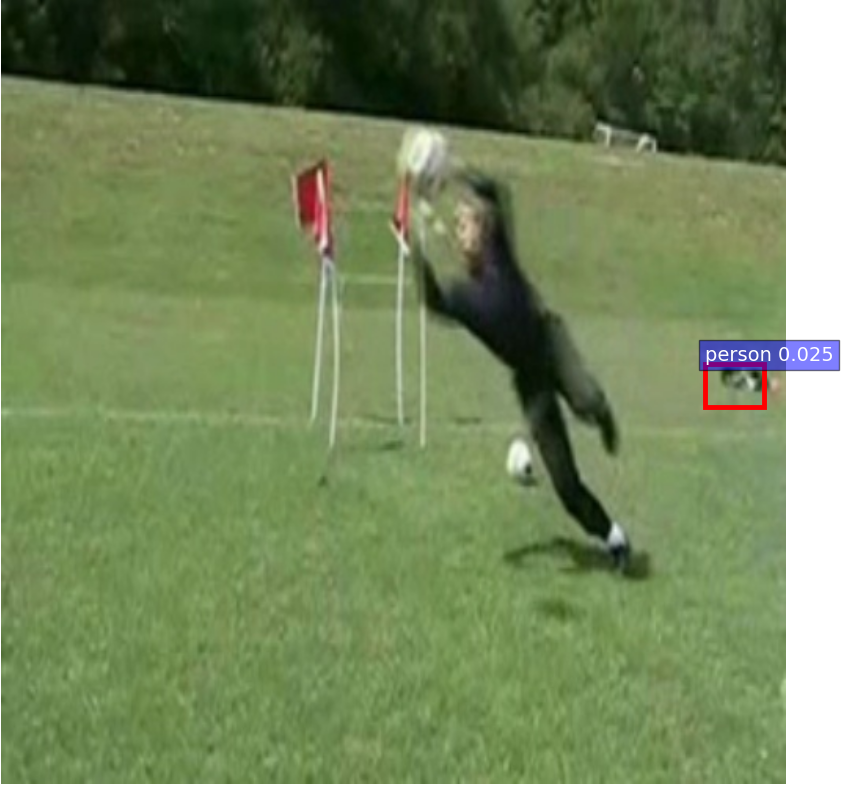}
  \caption{}
  \label{fig:sub1}
\end{subfigure}%
\hfill
\begin{subfigure}{.1\textwidth}
\centering
  \includegraphics[height = 0.85in]{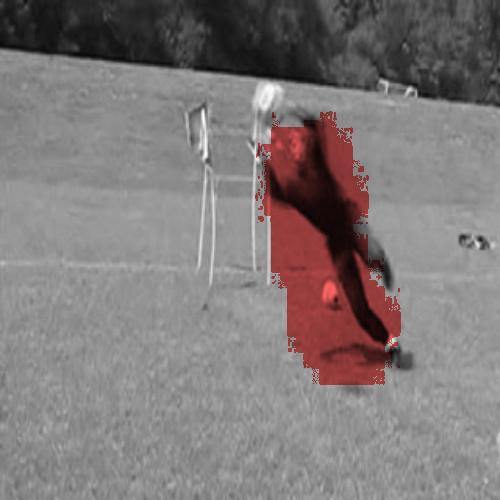}
  \caption{}
  \label{fig:sub1}
\end{subfigure}

\begin{subfigure}{.1\textwidth}
\centering
  \includegraphics[height = 0.85in]{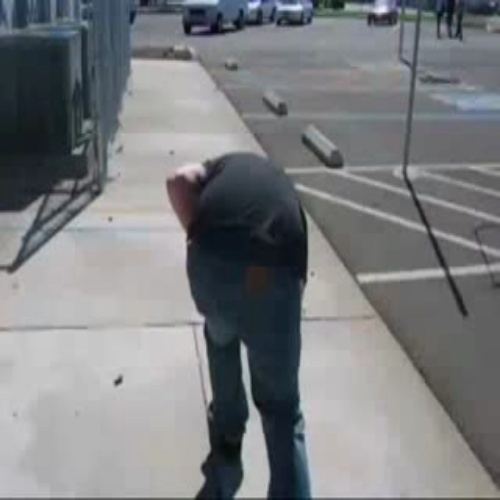}
  \caption{}
  \label{fig:sub1}
\end{subfigure}%
\hfill
\begin{subfigure}{.1\textwidth}
\centering
  \includegraphics[height = 0.85in]{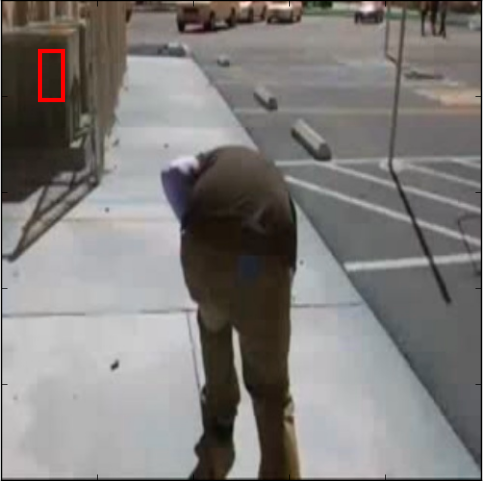}
  \caption{}
  \label{fig:sub1}
\end{subfigure}%
\hfill
\begin{subfigure}{.1\textwidth}
\centering
  \includegraphics[height = 0.85in]{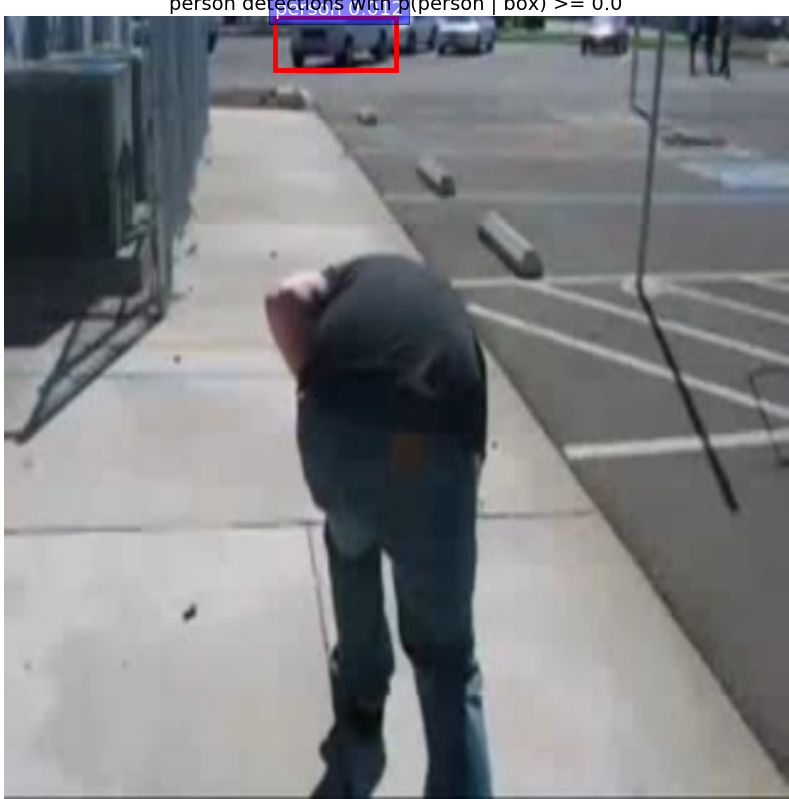}
  \caption{}
  \label{fig:sub1}
\end{subfigure}%
\hfill
\begin{subfigure}{.1\textwidth}
\centering
  \includegraphics[height = 0.85in]{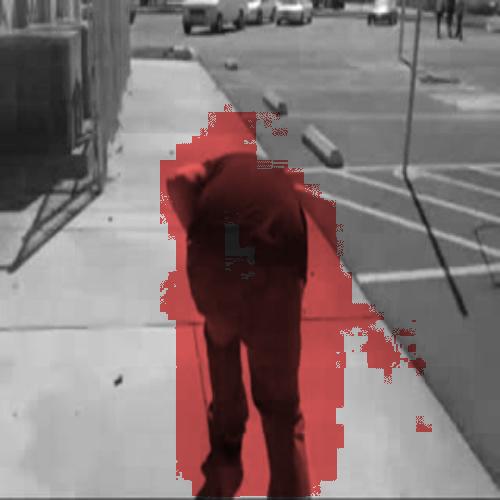}
  \caption{}
  \label{fig:sub1}
\end{subfigure}
\caption{Figure provides comparison of results with respect to supervised object detectors for person localisation. Figures (a),(e) and (i) are the original images, (b),(f) and (j) are results using the DPM \cite{dpm}, (c),(g) and (k) are results from Faster-RCNN and (d),(h) and (l) are from the proposed method. As can be seen, the automatically supervised method provides much better results even in hard examples not detected by the supervised object detectors.}
\label{fig:comparison}
\end{figure}

In figure~\ref{fig:comparison} we provide a comparison of the proposed method qualitatively as a localisation method against fully supervised method of Faster RCNN~\cite{faster_rcnn} that is a benchmark method for object localisation and deformable part model (DPM) approach \cite{dpm} that we use in our method as a means of person identification for various images. As can be seen from the figure, both the supervised object localisation methods fail to localise the person. This can be explained as the JHMDB dataset for action recognition \cite{jhmdb} has a different distribution of objects and the persons in figures~\ref{fig:comparison}(e) and (i) are not in a usual upright pose. However, the proposed method succeeds in estimating the pose of the persons accurately though the proposed method has not seen a single image from the JHMDB dataset during training. This shows the efficacy of the method in being able to localise persons accurately and even performing much better than the base method it was trained on.

\begin{figure*}
\centering
\begin{subfigure}{.19\textwidth}
\centering
  \includegraphics[height = 1.38in]{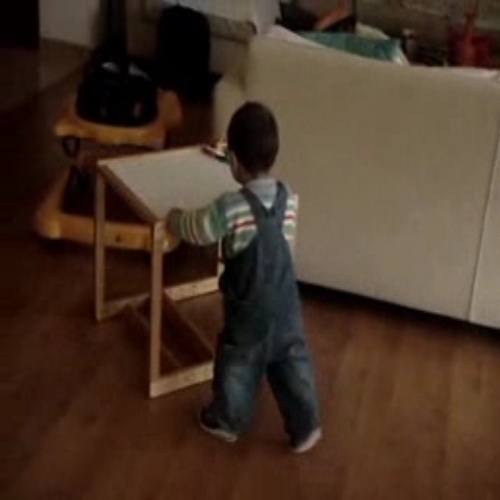}
  \caption{}
  \label{fig:sub1}
\end{subfigure}%
\hfill
\begin{subfigure}{.19\textwidth}
\centering
  \includegraphics[height=1.38in]{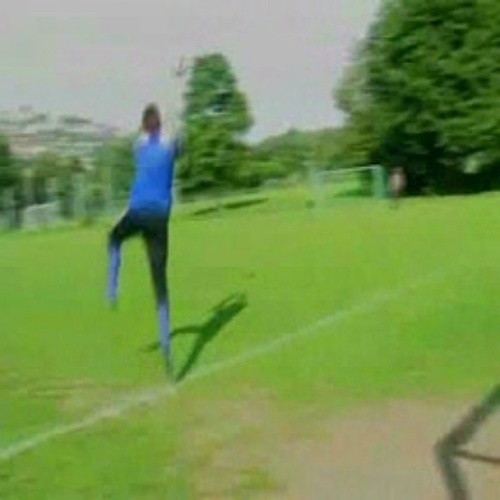}
  \caption{}
  \label{fig:sub2}
\end{subfigure}%
\hfill
\begin{subfigure}{.19\textwidth}
\centering
  \includegraphics[height = 1.38in]{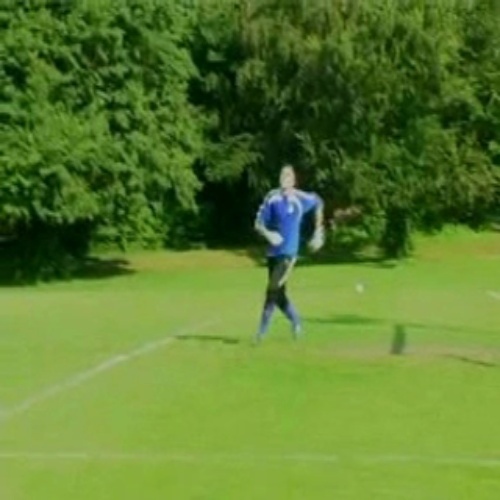}
  \caption{}
  \label{fig:sub1}
\end{subfigure}%
\hfill
\begin{subfigure}{.19\textwidth}
\centering
  \includegraphics[height=1.38in]{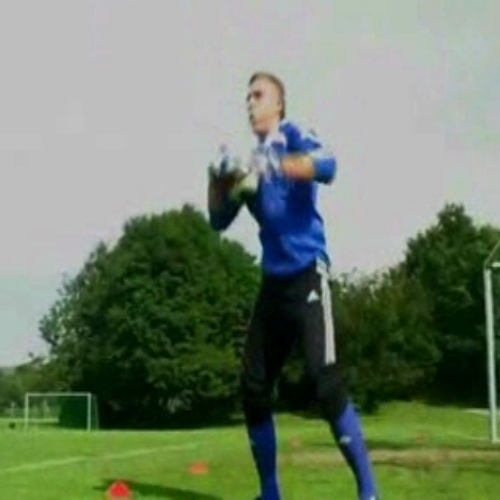}
  \caption{}
  \label{fig:sub2}
\end{subfigure}%
\hfill
\begin{subfigure}{.19\textwidth}
\centering
  \includegraphics[height=1.38in]{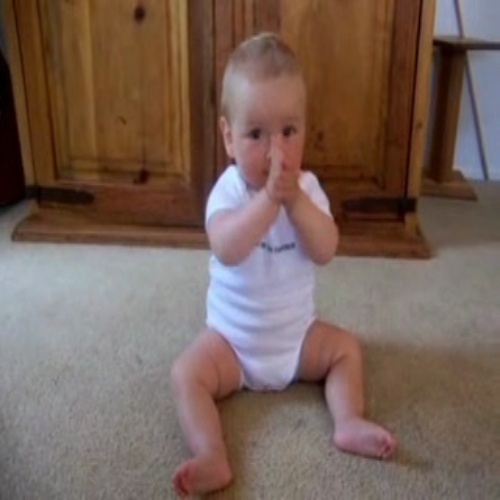}
  \caption{}
  \label{fig:sub2}
\end{subfigure}

\centering
\begin{subfigure}{.19\textwidth}
\centering
  \includegraphics[height = 1.38in]{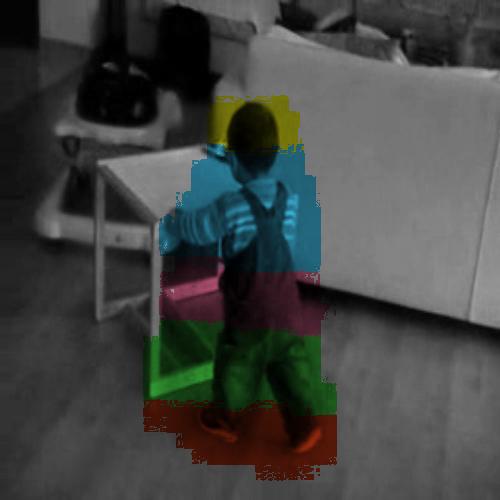}
  \caption{}
  \label{fig:sub1}
\end{subfigure}%
\hfill
\begin{subfigure}{.19\textwidth}
\centering
  \includegraphics[height=1.38in]{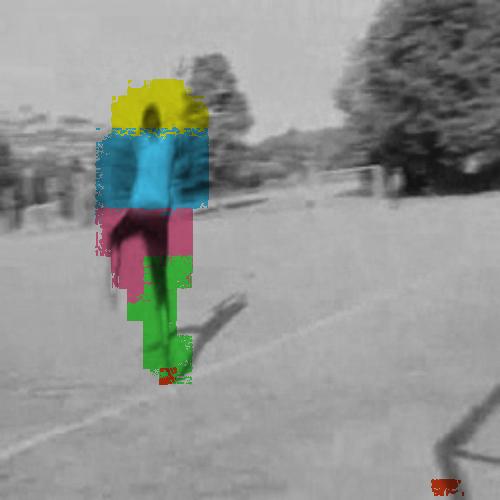}
  \caption{}
  \label{fig:sub2}
\end{subfigure}%
\hfill
\begin{subfigure}{.19\textwidth}
\centering
  \includegraphics[height = 1.38in]{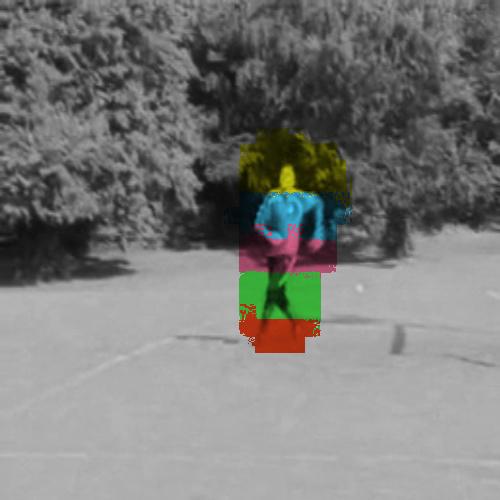}
  \caption{}
  \label{fig:sub1}
\end{subfigure}%
\hfill
\begin{subfigure}{.19\textwidth}
\centering
  \includegraphics[height=1.38in]{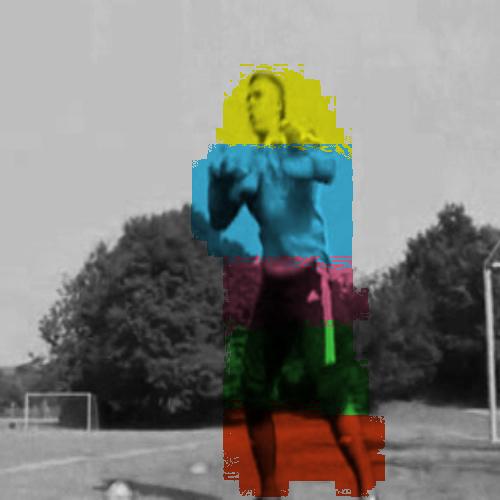}
  \caption{}
  \label{fig:sub2}
\end{subfigure}%
\hfill
\begin{subfigure}{.19\textwidth}
\centering
  \includegraphics[height=1.38in]{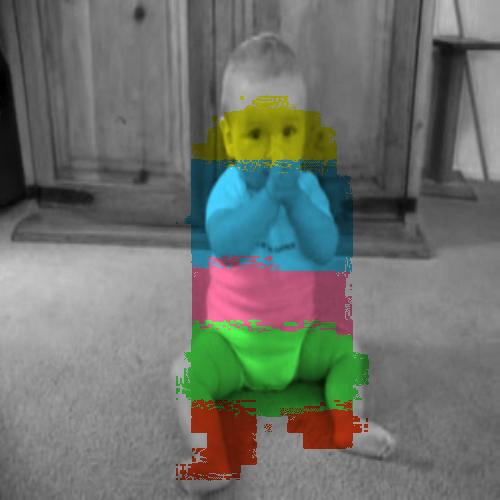}
  \caption{}
  \label{fig:sub2}
\end{subfigure}

\centering
\begin{subfigure}{.19\textwidth}
\centering
  \includegraphics[height = 1.38in]{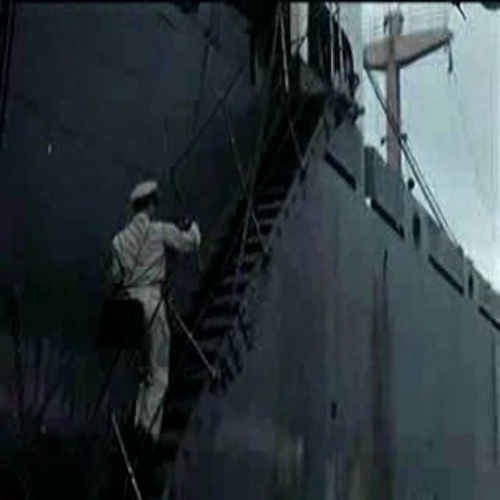}
  \caption{}
  \label{fig:sub1}
\end{subfigure}%
\hfill
\begin{subfigure}{.19\textwidth}
\centering
  \includegraphics[height=1.38in]{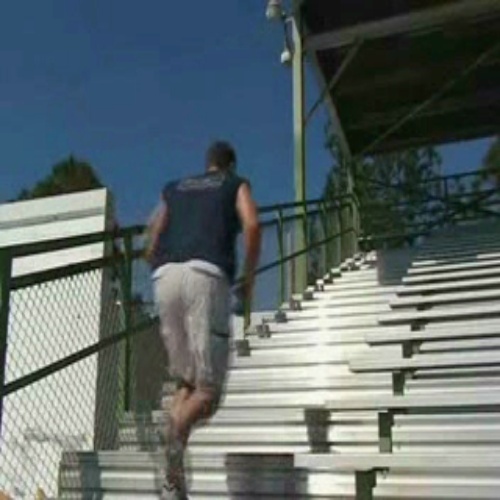}
  \caption{}
  \label{fig:sub2}
\end{subfigure}%
\hfill
\begin{subfigure}{.19\textwidth}
\centering
  \includegraphics[height = 1.38in]{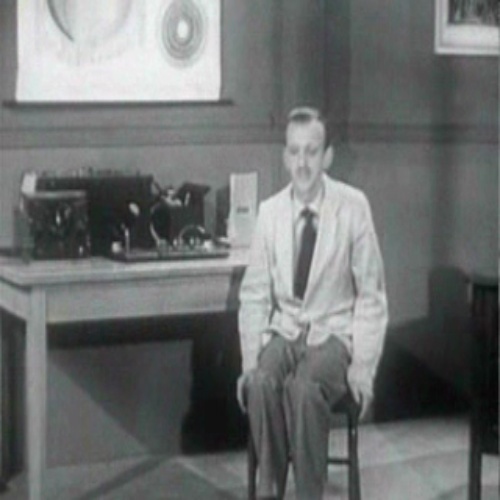}
  \caption{}
  \label{fig:sub1}
\end{subfigure}%
\hfill
\begin{subfigure}{.19\textwidth}
\centering
  \includegraphics[height=1.38in]{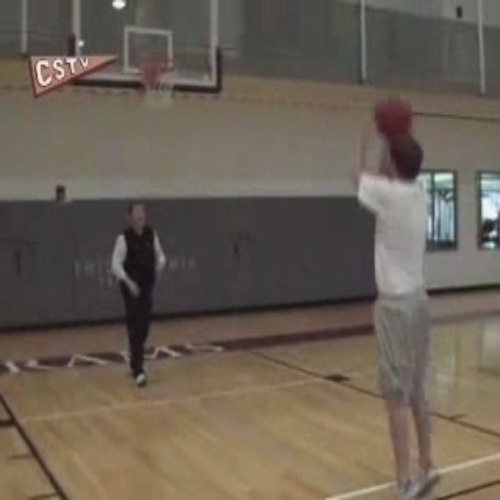}
  \caption{}
  \label{fig:sub2}
\end{subfigure}
\hfill
\begin{subfigure}{.19\textwidth}
\centering
  \includegraphics[height=1.38in]{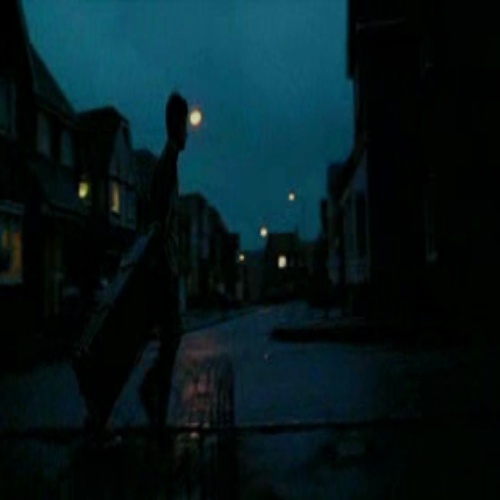}
  \caption{}
  \label{fig:sub2}
\end{subfigure}

\centering
\begin{subfigure}{.19\textwidth}
\centering
  \includegraphics[height = 1.38in]{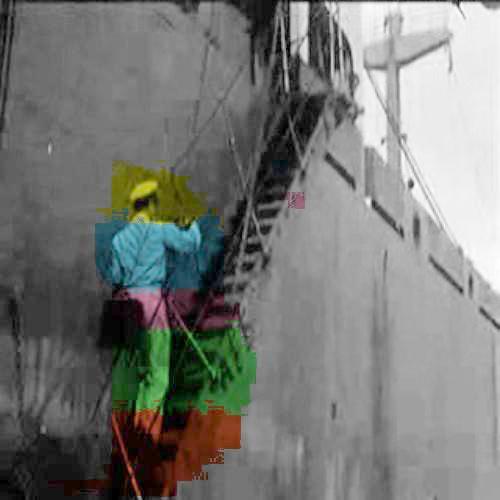}
  \caption{}
  \label{fig:sub1}
\end{subfigure}%
\hfill
\begin{subfigure}{.19\textwidth}
\centering
  \includegraphics[height=1.38in]{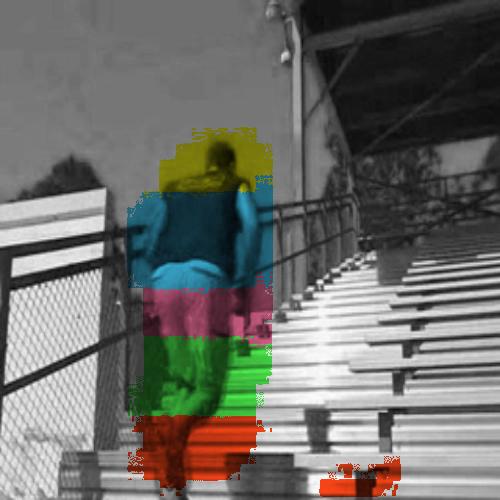}
  \caption{}
  \label{fig:sub2}
\end{subfigure}%
\hfill
\begin{subfigure}{.19\textwidth}
\centering
  \includegraphics[height = 1.38in]{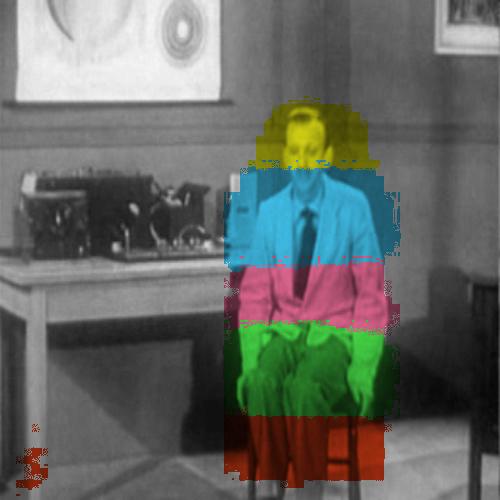}
  \caption{}
  \label{fig:sub1}
\end{subfigure}%
\hfill
\begin{subfigure}{.19\textwidth}
\centering
  \includegraphics[height=1.38in]{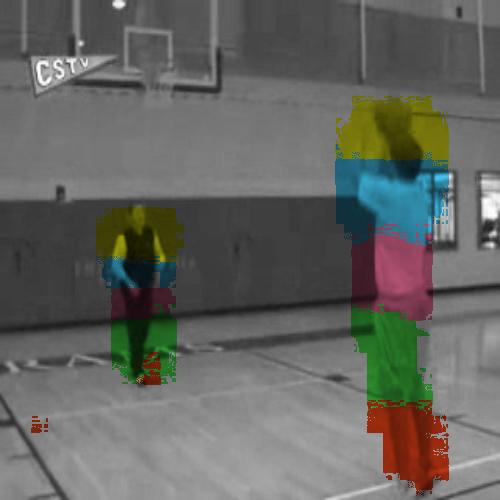}
  \caption{}
  \label{fig:sub2}
\end{subfigure}%
\hfill
\begin{subfigure}{.19\textwidth}
\centering
  \includegraphics[height=1.38in]{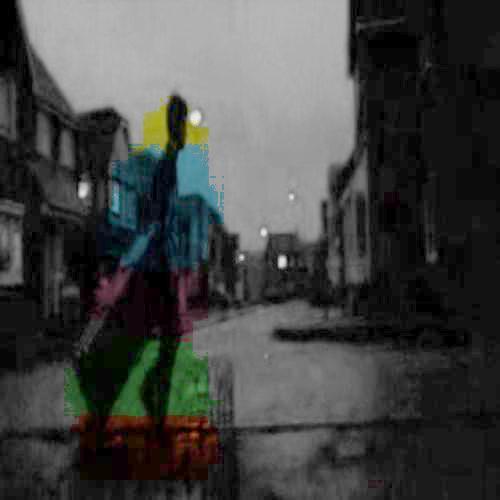}
  \caption{}
  \label{fig:sub2}
\end{subfigure}
\caption{Illustration of results: Figures (a) - (e) and (k) - (o) show the original images and (f) - (j)  and (p) - (t) show the respective pose estimates with the various colours depicting the different body parts estimated}
\label{fig:results}
\end{figure*}

As can be seen in figure~\ref{fig:results} the method performs very well on varying kinds of data ranging from complex pose of a child pushing a table (figure~\ref{fig:results}(a)and(f)) and a baby sitting ((figure~\ref{fig:results}(e)and(j)) to that of persons playing in the field (figures~\ref{fig:results}(b),(c),(d) and(g)(h)(i) ) to persons climbing stairs  (figure~\ref{fig:results}(l)and(q)) or ladder of a ship in adverse lighting (in this result figure~\ref{fig:results}(k) was the original image and the result figure~\ref{fig:results}(p) is enhanced for visualisation). Similarly figure~\ref{fig:results}(o) shows a person walking in the street at night and we show the result in figure~\ref{fig:results}(t) with enhanced brightness to visualise the result. Interestingly figure~\ref{fig:results}(m) shows a person sitting that is also accurately estimated as shown in figure~\ref{fig:results}{r}. Further figure~\ref{fig:results}(n) shows the generalization of the method towards estimating the pose of two people that are quite accurately estimated as shown in figure~\ref{fig:results}(s). Thus as can be seen the proposed method is applicable for a variety of images and provides us with a rather good estimate of pixel-wise dense pose estimates, albeit with fewer detail in terms of the exact joint locations.

\section{Conclusion}
\label{sec:conc}
We have obtained through this paper a method that can be automatically trained using basic techniques to obtain pose estimation from a single image without requiring any manual supervision. This is possible by harvesting data regarding coarse pose through the relative motion of people in videos. This method can be easily applied in various scenarios and shows robust dense pixel-wise estimates of human body pose in challenging situations. 

The limitation of the proposed method is in terms of being limited to only coarse blob based pose estimation. In future we would like to consider further advanced models such as hierarchical estimation of parts in order to obtain a more fine-grained pose for humans. To conclude, the performance of the proposed method without manual supervision is definitely encouraging and motivates the use of such self supervision for more tasks.
\clearpage 
{
\bibliographystyle{ieee}
\bibliography{egbib}
}

\end{document}